\documentclass{fairastra}
\usepackage[T1]{fontenc}
\usepackage{lmodern}
\usepackage{multirow}
\usepackage{soul}
\usepackage{pifont}
\usepackage[T1]{fontenc}

\usepackage{graphicx}
\usepackage{amsmath,amssymb}

\usepackage{makecell}
\usepackage[dvipsnames]{xcolor}
\usepackage{color, colortbl}
\usepackage[normalem]{ulem}
\usepackage{caption}
\usepackage{amssymb}
\usepackage{algorithm}
\crefname{table}{Table}{Tables}
\crefname{figure}{Figure}{Figrues}
\usepackage{textcomp}
\usepackage{marvosym}
\newcommand{\emailsym}{\Letter}

\usepackage{xspace}
\makeatletter
\DeclareRobustCommand\onedot{\futurelet\@let@token\@onedot}
\def\@onedot{\ifx\@let@token.\else.\null\fi\xspace}

\makeatother

\usepackage{newtxtext}

\usepackage{algpseudocode}
\definecolor{catgray}{gray}{0.92}

\usepackage[dvipsnames]{xcolor} 
\definecolor{FutureOrange}{HTML}{EC866D}

\title{AstraNav-Memory: Contexts Compression for Long Memory}

\author[1, 2*]{Botao Ren}
\author[1*\dagger\raisebox{-3pt}{\textsuperscript{\emailsym}}]{Junjun Hu}
\author[1, 3*]{Xinda Xue}
\author[1]{Minghua Luo}
\author[1, 3]{Jintao Chen}
\author[1]{Haochen Bai}
\author[3]{Liangliang You}
\author[1]{Mu Xu}

\affiliation[1]{\textbf{Amap, Alibaba Group}}
\affiliation[2]{\textbf{Tsinghua University}}
\affiliation[3]{\textbf{Peking University}}

\contribution[*]{Equal Contribution}
\contribution[\dagger]{Project Lead}
\contribution[]{\raisebox{-1pt}{\textsuperscript{\emailsym}}Corresponding authors}

\abstract{
\begin{abstract}

Lifelong embodied navigation requires agents to accumulate, retain, and exploit spatial–semantic experience across tasks, enabling efficient exploration in novel environments and rapid goal reaching in familiar ones. While object-centric memory is interpretable, it depends on detection and reconstruction pipelines that limit robustness and scalability. We propose an image-centric memory framework that achieves long-term implicit memory via an efficient visual context compression module end-to-end coupled with a Qwen2.5-VL–based navigation policy. 
Built atop a ViT backbone with frozen DINOv3 features and lightweight PixelUnshuffle+Conv blocks, our visual tokenizer supports configurable compression rates; for example, under a representative 16$\times$ compression setting, each image is encoded with about 30 tokens, expanding the effective context capacity from tens to hundreds of images.
Experimental results on GOAT-Bench and HM3D-OVON show that our method achieves state-of-the-art navigation performance, improving exploration in unfamiliar environments and shortening paths in familiar ones. Ablation studies further reveal that moderate compression provides the best balance between efficiency and accuracy. These findings position compressed image-centric memory as a practical and scalable interface for lifelong embodied agents, enabling them to reason over long visual histories and navigate with human-like efficiency.
\end{abstract}
}

\date{December 25th, 2025}

\astradata[Project Page]{\url{https://astra-amap.github.io/AstraNav-Memory.github.io/}}

\begin{document}

\maketitle

\section{Introduction}
With the recent progress in embodied navigation, the research focus has been shifting towards more complex multi-task navigation settings, giving rise to a series of new benchmarks~\cite{song2025towards,zhou2024navgpt,hu2024hiagent,zhang2024vision}. In real applications (e.g., household assistance), navigation is often lifelong: agents carry memory across tasks, gradually exploring and inferring the next subgoal in unfamiliar environments~\cite{yokoyama2024hm3d,xue2025omninav}, while leveraging prior experience to reach targets quickly in familiar environments~\cite{khanna2024goat,wang2024jarvis}. This mirrors human navigation—explore and reason upon first arrival in a new scene, then follow optimal paths based on memory once the environment is known~\cite{verma2016investigating,epstein2017cognitive}. The key to lifelong navigation lies in building effective spatial and semantic memory, enabling long-term visual histories to be efficiently stored, retrieved, and transformed into navigational advantages.

In embodied navigation, images are the most direct input and the fundamental carrier of memory. Research around what to remember and how to remember has largely converged on two routes: object-centric memory and image-centric memory. Object-centric memory can be explicit or implicit: explicit methods~\cite{zhou2025fsr,armeni20193d,rosinol20203d,yang20253d} rely on reconstructions and semantic annotations to recover object coordinates and categories, building queryable semantic maps to support downstream planning; implicit methods like MTU3D~\cite{zhu2025move} store historical object semantics and states via sparse, vectorized object queries, avoiding full reconstruction. These object-centric approaches facilitate retrieval and offer strong interpretability, but heavily depend on upstream detection/segmentation, involve complex pipelines with coupled errors, and have limited cross-domain generalization.

By contrast, image-centric memory is a more end-to-end implicit paradigm: it preserves camera poses and multi-view images, allowing the model to learn spatial structure and semantic distributions internally~\cite{zhang2024uni,anwar2025remembr,chiang2024mobility}. This approach naturally aligns with unified training objectives and navigation policies, reducing bottlenecks and error propagation from external modules. The core challenge is long-term memory:to truly benefit agents in lifelong tasks, the model must retain hundreds or thousands of historical frames within the context. However, the raw visual stream is inherently filled with significant spatial and temporal redundancy , providing a natural basis for compression. Without strong visual compression, such lengthy context becomes prohibitively expensive in both computation and storage, and attention mechanisms can be overwhelmed by noise and distractions, making it hard to focus on key information ~\cite{hu20253dllm,shen2024longvu}. Therefore, visual context compression is critical to enabling long-term image-centric memory.

Recent progress in visual compression has been rapid: in embodied navigation, existing work applies streaming modeling and token selection/merging~\cite{wei2025streamvln,zhang2024navid,cheng2024navila} to alleviate context length; in general vision-language models, various token compression and structured pooling schemes have emerged~\cite{chen2024image,zhang2024sparsevlm,cao2023pumer}. Especially in the OCR domain, the latest method, DeepSeek-OCR~\cite{wei2025deepseek}, demonstrates industrial-grade efficiency: by leveraging windowed attention, highly compressed convolutional features, and Mixture-of-Experts (MoE) decoding, it compresses dense image patches into very few context tokens with only minimal semantic loss. Inspired by this, we introduce visual context compression into long-term memory for embodied navigation, aiming to support longer histories under higher compression ratios while stably retaining retrievable spatial and semantic information within implicit representations.

We propose an image-centric memory framework centered on an efficient visual context compression module, end-to-end coupled with the navigation policy. Specifically, we build a structured compression network on top of Qwen2.5-VL’s native ViT, achieving approximately 20× token compression. Concretely, in our experimental setup, a $720\times 640$ RGB observation is tokenized into 598 visual tokens by the native ViT, while our two-stage compression reduces it to around 30 tokens. This drastic reduction transforms the usable context budget, enabling the agent to scale from storing only tens of images to maintaining hundreds of historical frames in context, thereby meeting the long-term implicit memory requirements of indoor scenes.

We evaluate on a lifelong navigation benchmark. In unknown environments, image-centric memory enables progressive reasoning, improving exploration efficiency and success rate; in familiar environments, long-term implicit memory shortens paths and reduces steps. Compared with explicit maps and implicit object queries, our image-centric memory offers advantages in end-to-end training, cross-domain robustness, and engineering simplicity; relative to existing image-centric methods, our compression significantly extends the maintainable history length and improves navigation metrics. Ablation studies show the clear impact of compression ratio and memory length; about 30 tokens per image achieves the best trade-off between efficiency and effectiveness while preserving spatial-semantic fidelity.
In brief, our work offers three primary contributions:
\begin{itemize}
\item We propose a unified framework for embodied lifelong navigation that uses an image-centric context memory to end-to-end couple vision, language, and decision-making, and incorporates an efficient visual context compression module that reduces the native ViT token count by ~20×, achieving high-fidelity representation with only ~30 tokens per frame, thereby accommodating hundreds of historical frames within a single context to enable large-scale, long-term implicit memory.

\item We propose a plug-and-play, ViT-native visual tokenizer: it passes frozen DINOv3 features through modules constructed of PixelUnshuffle and Convolution, and feeds them directly into the first block of Qwen2.5-VL-3B ViT without changing later modules. This design greatly lowers long-horizon cost while preserving mid-level spatial cues, and serves as a general-purpose compression module for navigation and other embodied tasks.

\item State-of-the-art performance on standard indoor embodied navigation benchmarks; ablations and analyses verify that the compressed implicit memory effectively encodes spatial and semantic information useful for planning; the benefits of long-term memory are particularly pronounced in familiar environments without sacrificing exploration in unfamiliar ones.
\end{itemize}
Overall, through strong visual compression and task-aligned training, image-centric memory enables implicit representations to replace object-centric mapping, providing more flexible and robust memory support for lifelong navigation. This direction is poised to become a unified memory interface for embodied agents, allowing them to continually accumulate experience, adapt rapidly, and move toward deployment via a more streamlined and reliable engineering path.
\begin{figure*}[!t]
    \centering
    \includegraphics[width=1\linewidth]{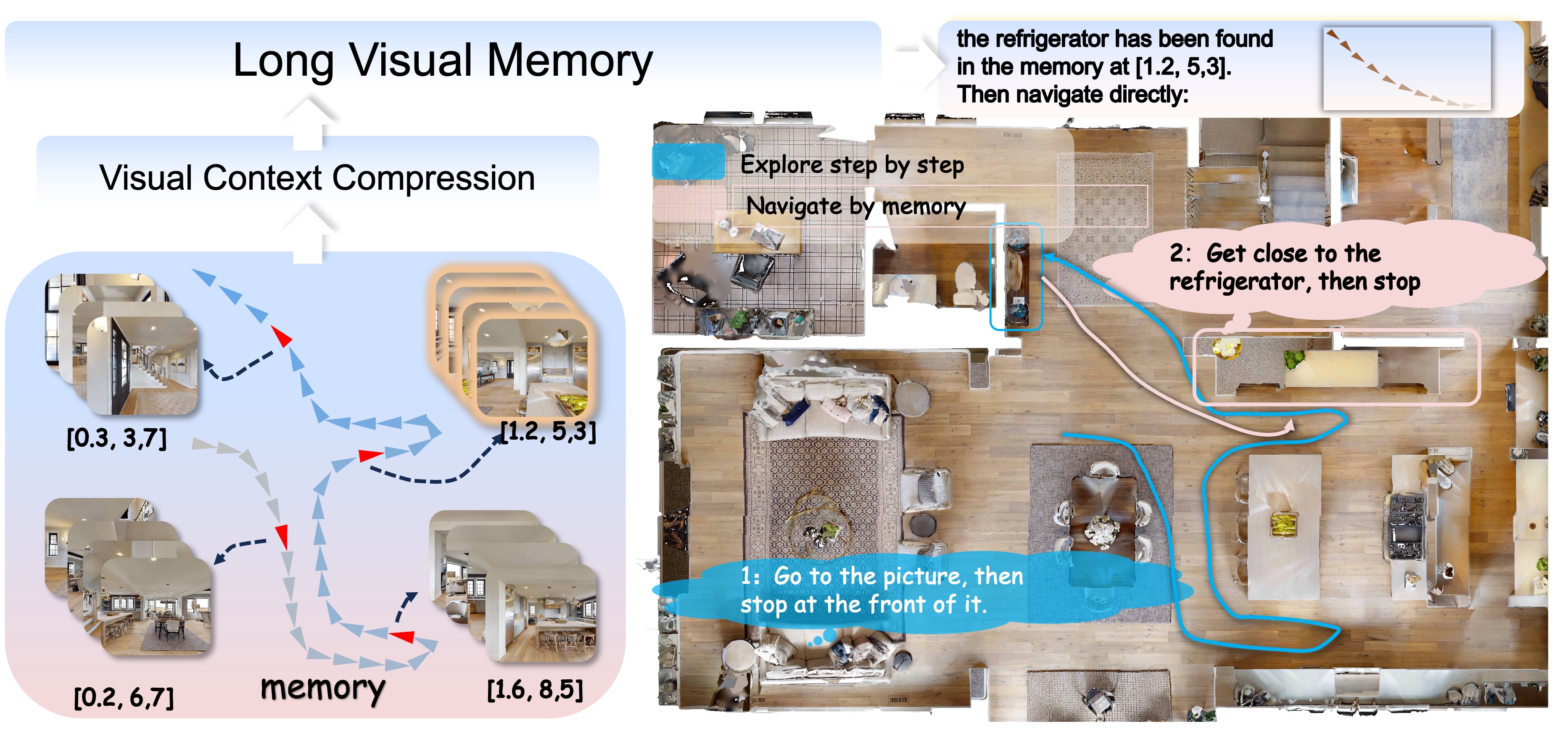}
    \caption{Our agent operates in a lifelong learning setting. For the initial task in an unseen environment, it uses frontier-based exploration to locate the target. Critically, the environment and agent state are preserved across tasks. For subsequent instructions, the agent first consults its memory. If the target object has been previously observed, the agent plans a direct path to its location, bypassing the need for re-exploration.}
    \label{fig:overview}
\end{figure*}

\section{Related Works}
\label{sec:formatting}

\subsection{Long-term memory}
Long-term memory has been extensively explored in LLMs/VLMs and embodied AI and is crucial for navigation; however, existing approaches still fall short in scalability, cross-domain robustness, and closed-loop stability. Broadly, prior work falls into two lines:
The first line is \textbf{explicit external memory with retrieval}.
These methods use memory stored outside the model and queried on demand via retrieval-augmented pipelines and structured indices. Foundational systems include RAG, REALM, kNN-LM, and RETRO, along with long-document and structured variants such as LongRAG, RAPTOR, GraphRAG, and OS-style MemGPT~\cite{lewis2020retrieval,guu2020retrieval,khandelwal2019generalization,borgeaud2022improving,jiang2024longrag,sarthi2024raptor,edge2024local,packer2023memgpt}. These methods offer interpretability and controllable access but relies on chunking and indexing heuristics, is sensitive to retriever recall and latency, and often fragments long-horizon temporal coherence, weakening coupling to closed-loop planning.
The second category is \textbf{implicit, token-integrated or parameterized memory}.
Memory is fused into the model’s states, tokens, or parameters through similarity-keyed memory banks and long-context modeling. Influential directions include context-window scaling (e.g., YaRN, LongRoPE2) and streaming/recurrent attention (e.g., StreamingLLM, Infini-Attention, RMT), extended to multimodal settings by Flamingo, LongVILA, LongVLM, LLaMA-VID, and Long-Context SSM Video World Models~\cite{munkhdalai2024leave,ding2024longrope,xiao2023efficient,bulatov2023scaling,alayrac2022flamingo,chen2024longvila,weng2024longvlm,li2024llama,po2025long}. A complementary thread enhances memory at test time via on-the-fly adaptation, from early TTT/Tent to recent Titans, which learn persistent memory during inference~\cite{sun2020test,wang2020tent,behrouz2024titans}. This line enables end-to-end training without external indices but still lacks structured retrievability over entities, locations, and temporal relations, incurs growing attention cost and noise with long histories, and faces capacity–staleness trade-offs in memory banks.

Existing methods still have inconsistent offline versus closed-loop evaluation, dependence on oracle perception or poses, weak coupling between retrievability and planning, and a mismatch between compression goals and control utility. Our approach falls under the implicit line and introduces task-aligned visual context compression that yields compact yet semantically retrievable visual tokens, reducing computation and storage, mitigating attention noise, and preserving spatial–semantic cues for long-horizon planning.


\subsection{Visual context compression and token efficiency}

Visual context compression seeks to reduce the number of visual tokens while preserving navigation-critical spatial–semantic information, thereby enabling longer contexts, lower latency, and broader deployment. Recent efforts have converged around three methodological categories.
1. Pre-encoding and merging. PVC, VScan, and InternVL-X reduce tokens inside the vision stack via hierarchical aggregation and global–local fusion, achieving sizable savings while retaining coarse layout~\cite{yang2025pvc,zhang2025vscan,lu2025internvl}. However, objectives are largely perception-driven rather than aligned with downstream control, and constraints on long-horizon consistency are limited.
2. Pruning and sparsification. TokenCarve, SparseVLM, and FocusLLaVA remove or down-weight low-information tokens at the encoder or within the language model; an empirical study reports instability and cases where pruning underperforms simple pooling~\cite{tan2025tokencarve,  zhang2024sparsevlm,zhu2024focusllava,wen2025token}. These methods are simple and plug-and-play but often fail to reliably retain key frames or viewpoints and struggle under distribution shift.
3. Model-internal and language-side compression. Some methods integrate compression with the language model or the end-to-end objective~\cite{xing2025vision,li2025fcot,ye2025voco}. DeepSeek-OCR demonstrates extreme compression with strong efficiency gains but it targets text recognition and does not guarantee preservation of spatial semantics, retrievability, or planning utility in embodied settings~\cite{wei2025deepseek}.
\begin{figure*}[t]
    \centering
    \includegraphics[width=0.99\linewidth]{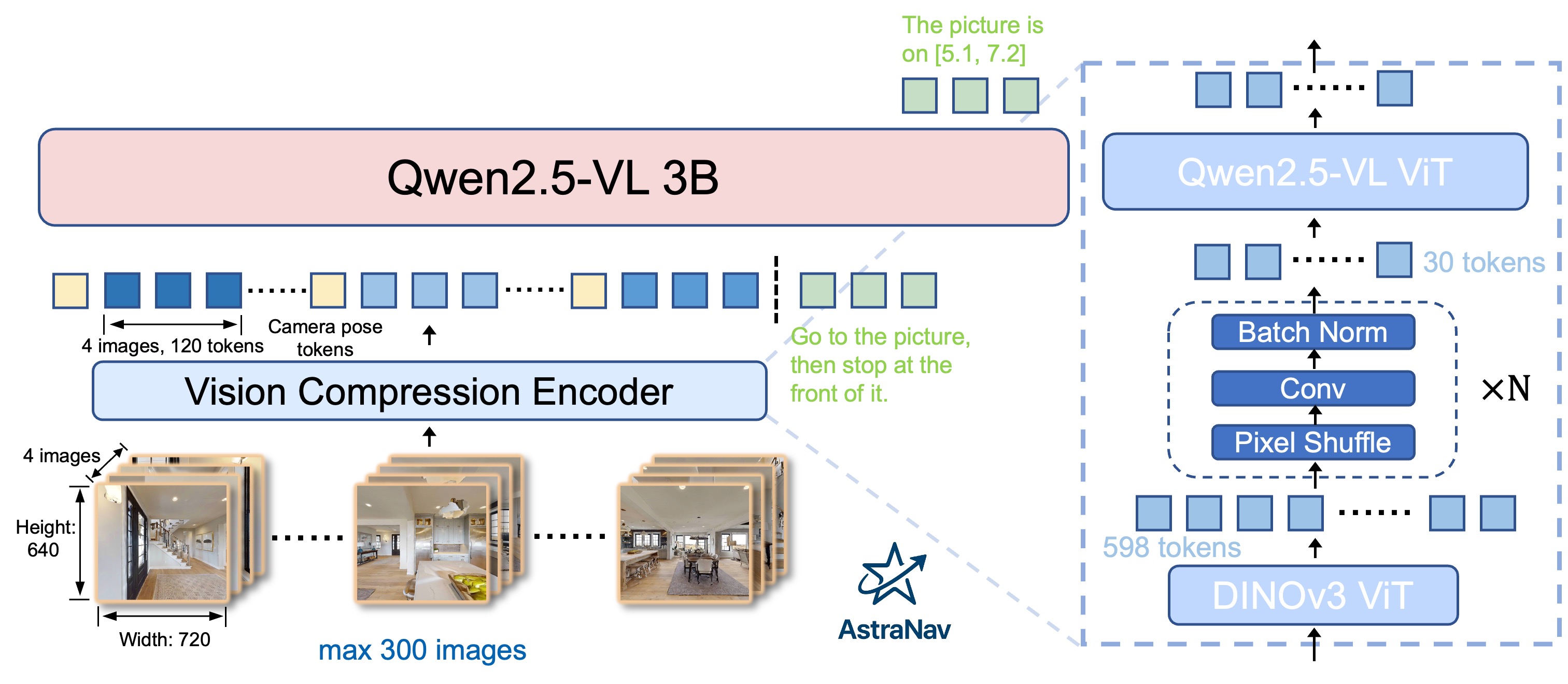}
    \caption{Overview of AstraNav-Memory with the proposed compressed vision encoder. During navigation, up to 300 images are first encoded by a DINOv3 ViT into 598 visual tokens, which are then compressed by several lightweight compression heads into 30 tokens compatible with the original Qwen2.5-VL ViT. The compact visual tokens and the language command are fed into Qwen2.5-VL-3B, enabling long-horizon navigation reasoning over large visual memories at low computational cost.}
    \label{fig:model_figure}
\end{figure*}

Above all, these three strands leave two persistent gaps: compression remains perception-driven with weak constraints on long-horizon consistency and structured retrievability, and compute plus attention noise grow unfavorably with history length. We address this with an image-centric, task-aligned compression objective trained end-to-end with the navigation policy. The resulting tokens deliver order-of-magnitude reduction while preserving planning-critical cues, enabling much longer visual histories under fixed budgets and reducing attention noise. In turn, we observe stronger closed-loop stability in novel scenes and greater efficiency in familiar ones.

\section{Method}

We present AstraNav-Memory, which compresses each frame into a compact token sequence and feeds it to the Qwen2.5-VL-3B ViT. It reduces the per-frame sequence from 598 tokens to 30. The language model remains unchanged.

As shown in \cref{fig:overview}. The model extract features with DINOv3-ViT-Base. Two PixelUnshuffle stages follow, each paired with a channel-alignment block. This sequence is flattened and passed directly to the first block of the Qwen2.5-VL-3B ViT. Then apply a 2$\times$2 patch-merger to form 30 tokens. The compressor preserves mid-level spatial cues—landmarks, layout, traversability—while substantially reducing sequence length.

\subsection{Preliminary}

Vanilla Qwen2.5-VL uses VisionPatchEmbed (patch=14) followed by the Qwen2.5-VL-3B ViT. The ViT outputs are then compressed by a built-in 2×2 patch-merger, projected through the multimodal projector, and finally fed into the LM.
We cast navigation as a sequence-to-sequence problem. The model input is
\begin{equation}
    \mathbf{x} = [\text{SYS}; (P_1, I_1); (P_2, I_2); \ldots; (P_T, I_T); \text{INSTR}],
\end{equation}
where \text{SYS} is the system prompt, \text{INSTR} is the current instruction, and $(P_t, I_t)$ denotes the $t$-th camera-pose and image pair. The camera pose $P_t$ is serialized as text tokens and inserted immediately before the visual tokens of $I_t$, so all pose information is processed by the model’s text encoder. The model then outputs natural language describing either a frontier or a target, whose 2D locations are wrapped in special tags, e.g., \texttt{<coordinate>x, y</coordinate>}.
As the history grows, the computational cost scales quadratically with the total number of tokens, i.e., $O((\text{tokens per frame} \times \text{history length})^2)$, leading to a severe long-horizon bottleneck.
We change only the visual input sequence. All later modules stay the same.

\subsection{Model Architecture}
\textbf{Base feature extractor.} We first use DINOv3-ViT-Base to extract features from 2D images of the 3D scene. DINOv3 is chosen for its strong self-supervised semantics, robustness to domain shifts, and ability to capture mid-level spatial cues without task labels. The representation is stable under lighting and texture changes and transfers well to unseen environments, which is crucial for VLN exploration.
The DINOv3 backbone is frozen to stabilize semantics, reduce training cost, and decouple the compressor and policy from the encoder, preventing co-adaptation and enabling drop-in backbone upgrades.

\textbf{Compression.} To reduce sequence length while retaining geometric information, we adopt rearrangement rather than pooling. PixelUnshuffle moves each local $2 \times 2$ neighborhood into channels,
\begin{equation}
    PU_2: \mathcal{R}^{H\times W\times C} \rightarrow \mathcal{R}^{\frac{H}{2} \times \frac{W}{2} \times (4C)}
\end{equation}

\textbf{Compression block.} After each PixelUnshuffle (stride 2), the tensor passes through a stride 1, $3 \times 3$ convolution followed by BatchNorm and SiLU. We refer to this sequence as a compression block. A single block halves the height and width, thus reducing the spatial token count by $4 \times$ while remapping channels:

\begin{equation}
    \mathbf{X}^{(i)} = \text{SiLU}(\text{BN}(\text{Conv}(\mathbf{X}^{(i-1)})))
\end{equation}

where $(H_i, W_i)=(H_{i-1}/2, W_{i-1}/2)$.

A stack of $N$ compression blocks yields a total spatial compression of

\begin{equation}
    r = 2^{2N} = 4^N, (H_N, W_N) = (\frac{H_0}{2^N}, \frac{W_0}{2^N})
\end{equation}

If a final $2 \times 2$ patch-merger is applied, $(H_N, W_N)$ must be even. We therefore pad the DINO feature patch grid so that $H_0, W_0$ are multiples of $2^{N+1}$. which ensures that both the downsampled grid and the merger are well-defined.

On the downsampled grid, we add 2D positional encodings. Let $\hat{\mathbf{X}}^{(N)}\in\mathcal{R}^{H_N\times W_N\times C_N}$ denote the output of the last block. After flattening and a $2\times2$ patch-merger $\mathcal{M}_2$, the per-frame sequence is

\begin{equation}
    \tilde{\mathbf{Z}}_t = \mathcal{M}_2(\mathrm{Flatten}(\hat{\mathbf{X}}^{(N)})) + \mathbf{P}^{(2\mathrm{D})}, \tilde{\mathbf{Z}}_t \in \mathcal{R}^{L_t\times C_N}
\end{equation}

In our main setting with two blocks $(N=2)$, a $720\times640$ input (DINO patch (16)) yields $H_0\times W_0=45\times40$, padded to $48\times40$, then $(H_2,W_2)=(12,10)$. After the merger, $L_t=\frac{12\cdot10}{4}=30$.

To interface with Qwen2.5-VL-3B ViT, the channel dimension $C_N$ is matched to the ViT hidden size (1280) by the last block (the stride 1, $3\times3$ projection in $\mathrm{Conv}(\cdot))$. The resulting tokens $\tilde{\mathbf{Z}}_t\in\mathcal{R}^{30\times1280}$ are fed directly into the first ViT block, without CLS or register tokens. The vision-language projector and the language model remain unchanged.
\subsection{Dataset Construction}

\textbf{Open-vocabulary object navigation data (OVON).}
We use the OVON~\cite{yokoyama2024hm3d} dataset and follow the data generation methodology proposed in MTU3D~\cite{zhu2025move}. Navigation data is comprised of four discrete actions: MOVE\_FORWARD, TURN\_LEFT, TURN\_RIGHT, and STOP. Each action corresponds to a continuous trajectory point, which is defined by its 6-DOF (Degrees of Freedom) pose. This pose consists of a 3D position $(X, Y, Z)$ and an orientation represented as a quaternion $(w, x, y, z)$. Then, for each training data, we randomly sample a pair of start pose and target object category within the scene. During the agent's exploration, we maintain a 3D occupancy map that classifies each region as either "explored" or "unknown". The frontier is defined as the boundary points between the two regions. The next-subgoal selection policy first computes the shortest-path cost from each candidate frontier to the target and ranks them. It then prefers the shortest path while introducing limited randomness: with high probability it selects the minimum-cost frontier, and with a small probability it randomly samples one from the remaining candidates, balancing efficiency and exploration diversity. A successful task, which terminates upon finding the target, is recorded as a tuple containing: a first-person RGB video stream, a sequence of frontier-based sub-goals that guided the exploration, and a descriptive natural language instruction. Ultimately, we constructed the OVON-500K training dataset by uniformly sampling from all 145 available scenes.

\textbf{Lifelong Navigation data (GOAT).} 
A novel and more challenging navigation task named GOAT-Bench~\cite{khanna2024goat} that integrates multi-modal goal navigation within a lifelong learning paradigm. Different from the episodic and single-goal settings of tasks like OVON, agent must continuously navigate to a sequence of target objects in a persistent indoor environment (HM3D) in GOAT-Bench. The defining characteristic of this task is its state continuity: upon completing a sub-task, neither the environment nor the agent's state is reset, instead, the agent builds upon a persistent memory of history experiences and executes its next instruction after it finishes current sub-task. This continuous setup is explicitly designed to evaluate the agent's ability to accumulate and leverage long-term spatial knowledge from prior exploration to perform next following sub-tasks with higher efficiency. Consequently, we designed several datasets from all 136 distinct scenes, each corresponding to a different memory length: 50, 100, 200, and 500 steps. This resulted in four datasets, namely GOAT-1M-50L, GOAT-1M-100L, GOAT-1M-200L, and GOAT-1M-500L. For each dataset, we subsequently performed a data filtering process to prevent an over-representation of overly short trajectories.

\section{Experiment}

\subsection{Experiment setting}
We fine-tune Qwen2.5-VL-3B as our base model. The training set contains 1.5M samples, obtained by combining OVON-500K with the GOAT-1M-50L/100L/200L/500L subsets. For experiments with different numbers of input images, we use the GOAT split whose history length (50/100/200/500 frames) matches the desired visual context length. Training is performed on 32 H20 GPUs with a learning rate of $1\times10^{-5}$, a warmup ratio of 0.05, and at most 2 epochs of optimization. We evaluate models using Success Rate (SR) and Success weighted by Path Length (SPL). A trajectory is counted as successful only if the final agent position is within 1 m of the target, so higher SR is better. SPL further measures how efficient successful trajectories are, and is computed as
\begin{equation}
    \text{SPL} = \frac{1}{N}\sum_{i=1}^{N} S_i \frac{L_i^{*}}{\max(L_i, L_i^{*})},
\end{equation}
where $S_i\in{0,1}$ denotes whether case (i) succeeds, $L_i^{*}$ is the shortest-path length to the goal, and $L_i$ is the actual path length. Both SR and SPL are higher-is-better metrics.

\begin{table*}[t]
  \centering
  \caption{Success Rate (SR) and Success weighted by Path Length (SPL) on GOAT-Bench for Multi-modal Lifelong Navigation.}
  \label{tab:goat}
  \setlength{\tabcolsep}{4pt} 

  \begin{tabular}{l|cc|cc|cc}
    \toprule
     & \multicolumn{2}{c|}{\textbf{Val-Seen}} & \multicolumn{2}{c|}{\textbf{Val-Seen-Synonyms}} & \multicolumn{2}{c}{\textbf{Val-Unseen}} \\
    \cmidrule(lr){2-3} \cmidrule(lr){4-5} \cmidrule(lr){6-7}
    \textbf{Method}& \textbf{SR}$\uparrow$ & \textbf{SPL}$\uparrow$ & \textbf{SR}$\uparrow$ & \textbf{SPL}$\uparrow$ & \textbf{SR}$\uparrow$ & \textbf{SPL}$\uparrow$ \\
    \midrule
    Modular GOAT ~\citep{chang2023goat}     & 26.3 & 17.5  & 33.8  & 24.4  & 24.9  & 17.2  \\
    Modular CLIP on Wheels ~\citep{gadre2023cows}  & 14.8 & 8.7  & 18.5  & 11.5  & 16.1  & 10.4  \\
    SenseAct-NN Skill Chain   ~\citep{khanna2024goat}     & 29.2 & 12.8  & 38.2 & 15.2  & 29.5 & 11.3  \\
    SenseAct-NN Monolithic  ~\citep{khanna2024goat}     & 16.8 & 9.4 & 18.5 & 10.1 & 12.3 & 6.8  \\
    TANGO~\citep{Ziliotto2024TANGOTE}     & - & - & - & - & 32.1 & 16.5 \\
    MTU3D~\citep{zhu2025move}     & 52.2 & 30.5 & 48.4 & 30.3 & 47.2 & 27.7 \\
        \midrule
            \textbf{AstraNav-Memory} & \textbf{65.5} & \textbf{49.0} & \textbf{66.8} & \textbf{54.7} & \textbf{62.7} & \textbf{56.9}  \\
    \bottomrule
  \end{tabular}
\end{table*}
\begin{table*}[htbp]
  \centering
  \caption{Object-goal navigation results on HM3D-OVON.}
  \label{tab:ovon}
  \begin{tabular}{l|cc|cc|cc}
    \toprule
    \textbf{Method} & \multicolumn{2}{c|}{\textbf{Val-Seen}} & \multicolumn{2}{c|}{\textbf{Val-Seen-Synonyms}} & \multicolumn{2}{c}{\textbf{Val-Unseen}} \\
    \cmidrule(lr){2-3} \cmidrule(lr){4-5} \cmidrule(lr){6-7}
    & \textbf{SR}$\uparrow$ & \textbf{SPL}$\uparrow$ & \textbf{SR}$\uparrow$ & \textbf{SPL}$\uparrow$ & \textbf{SR}$\uparrow$ & \textbf{SPL}$\uparrow$ \\
    \midrule
    BC~\citep{pomerleau1988alvinn}         & 11.1 & 4.5  & 9.9  & 3.8  & 5.4  & 1.9  \\
    DAgger~\citep{ross2011reduction}     & 11.1 & 4.5  & 9.9  & 3.8  & 5.4  & 1.9  \\
    RL~\citep{schulman2017proximal}          & 18.1 & 9.4  & 15.0 & 7.4  & 10.2 & 4.7  \\
    DAgRL~\citep{chen2019touchdown}       & 41.3 & 21.2 & 29.4 & 14.4 & 18.3 & 7.9  \\
    BCRL~\citep{wang2019reinforced}         & 39.2 & 18.7 & 27.8 & 11.7 & 18.6 & 7.5  \\
    VLFM~\citep{yokoyama2024vlfm}        & 35.2 & 18.6 & 32.4 & 17.3 & 35.2 & 19.6 \\
    DAgRL+OD~\citep{yokoyama2024hm3d}     & 38.5 & 21.1 & 39.0 & 21.4 & 37.1 & 19.8 \\
    Uni-NaVid~\citep{zhang2024uni}   & 41.3 & 21.1 & 43.9 & 21.8 & 39.5 & 19.8 \\
        TANGO~\citep{Ziliotto2024TANGOTE}     & - & - & - & - & 35.5 & 19.5 \\
    MTU3D~\citep{zhu2025move}     & 55.0 & 23.6 & 45.0 & 14.7 & 40.8 & 12.1 \\

        \midrule
    \textbf{AstraNav-Memory} & \textbf{65.6} & \textbf{35.4} & \textbf{57.5} & \textbf{33.0} & \textbf{62.5} & \textbf{34.9} \\
    \bottomrule
  \end{tabular}
\end{table*}

\begin{table}[t]
\centering
\caption{Efficiency comparison for different numbers of stored images with and without 16$\times$ token compression.}
\label{tab:main_50_100}
\begin{tabular}{l|cccc}
\toprule
\# Images   & Acc. & \begin{tabular}[c]{@{}c@{}}Train time\\ (per iter/s)\end{tabular} & \begin{tabular}[c]{@{}c@{}}Mem.\\ (GB)\end{tabular} & \begin{tabular}[c]{@{}c@{}}Inf. time\\ (per instr./s)\end{tabular} \\ \hline
50 (origin) & 60.2 & 26.4                                                              & 90.6                                                &     10.3                                                     \\
50 (16$\times$)    & 56.6 & 6.5                                                               & 46.8                                                &    2.2                                                      \\
100 (16$\times$)   & 57.5 & 9.2                                                               & 68.5                                                &   4.2                                                       \\
200 (16$\times$)   & 55.2 & 12.0                                                              & 86.8                                                &    5.6                                                      \\
\bottomrule
\end{tabular}
\end{table}
\begin{table}[ht]
\centering
\caption{Evaluation of navigation accuracy on a \textbf{uniformly sampled subset} of Val-Unseen on GOAT-Bench with varying token compression rates and numbers of stored images.}
\label{tab:compression_rate}
\begin{tabular}{c|llll}
\toprule
Compression rate & 50   & 100  & 200  & 500 \\ \hline
1                & 60.2 & -    & -    & -   \\
4                & 61.3 & 63.2 & -    & -   \\
16               & 56.6 & 57.5 & 55.2 & -   \\
64               & 42.5 &  49.1    & 48.1 &  47.6   \\
\bottomrule
\end{tabular}
\end{table}
\begin{table}[ht]
\centering
\caption{Navigation success rate and parameter counts for different DINOv3 ViT backbones used in our visual tokenizer. The compression rate is 16$\times$.}
\begin{tabular}{c|cc}
\toprule
DINO ViT & SR & Param. \\ \hline
ViT-S    & 53.3 & 29M    \\
ViT-B    & 57.5 & 86M    \\
ViT-L    & 59.4 & 300M   \\ \bottomrule
\end{tabular}
\label{tab:dino_backbone}
\end{table}

\subsection{Quantitative result}
We conduct multi-modal lifelong navigation results on GOAT-Bench. We compare SR and SPL on the Val-Seen, Val-Seen-Synonyms, and Val-Unseen splits. Among existing approaches, as shown in \cref{tab:goat}, MTU3D is the strongest baseline, obtaining 52.2\% SR / 30.5\% SPL on Val-Seen and 48.4\% SR / 30.3\% SPL on Val-Seen-Synonyms, clearly outperforming modular pipelines such as Modular GOAT and Modular CLIP on Wheels. On the most challenging Val-Unseen split, our method achieves 62.7\% SR and 56.9\% SPL, significantly surpassing all prior methods. In particular, we improve over the previous state-of-the-art MTU3D (47.2\% SR, 27.7\% SPL) by +15.5\% SR and +29.2\% SPL, and achieve more than 2.4$\times$ higher SR and 3.2$\times$ higher SPL than Modular GOAT. These substantial gains demonstrate the strong generalization ability of our approach to unseen environments and instructions in the lifelong navigation setting.

\cref{tab:ovon} summarizes open-vocabulary navigation results on HM3D-OVON. Existing methods obtain moderate performance on the unseen split: behavior cloning (BC), DAgger, and standard RL achieve SR below 20\%, while more advanced approaches such as VLFM, DAgRL+OD, Uni-NaVid, and MTU3D improve SR to the 35$\sim$41\% range with SPL around 12$\sim$20\%. In contrast, our method attains 62.5\% SR and 34.8\% SPL, outperforming the previous best MTU3D by +21.7\% SR and +22.8\% SPL points. This corresponds to roughly 1.5$\times$ higher success rate and 1.7$\times$ higher path efficiency, indicating that our approach generalizes substantially better to unseen open-vocabulary navigation instructions.

\subsection{Ablation study}
\textbf{Effect of memory length under fixed compression.}
We conduct an ablation on memory length under a fixed compression setting. As shown in \cref{tab:main_50_100}, "50 (origin)" corresponds to the baseline that keeps 50 uncompressed images, while “N (16$\times$)” denotes using N input images, each compressed by a factor of 16 in token length. As shown, compression greatly improves efficiency: using 50 compressed images reduces training time per iteration from 26.4 s to 6.5 s and almost halves GPU memory (90.6 GB → 46.8 GB), with a similar trend for inference time. In terms of accuracy, 50 (origin) still performs the best, indicating that aggressive compression introduces a small performance drop. Among the compressed variants, 100 (16$\times$) $>$ 200 (16$\times$) $>$ 50 (16$\times$). We hypothesize that 100 (16$\times$) offers the best balance between history length and context size: compared with 50 (16$\times$), it can leverage more past observations, while 200 (16$\times$) suffers from an overly long context window, where the model has difficulty attending to the most recent—and often most informative—images.

\textbf{Effect of compression rate on performance.}
\cref{tab:compression_rate} studies the interaction between the compression rate and the number of stored images. As the compression rate increases, the model can accommodate a longer visual history under the same memory budget (e.g., 4$\times$ compression allows us to move from 50 to 100 images, and 16$\times$ further extends the length to 200 images). However, very aggressive compression is harmful: at 64$\times$, the performance drops sharply (42.5\% for 50 images and 48.1\% for 200 images), suggesting severe information loss in the compressed tokens. In practice, we find that 16$\times$ compression offers a good trade-off between training speed and the amount of observable history, while 4$\times$ compression with 100 images yields the best overall performance. Note that the maximum number of images is not simply proportional to the compression factor—for example, 1$\times$ with 50 images does not scale to 800 images at 16—because the DINOv3 ViT encoder must still keep its pre-compression tokens in memory, which dominate GPU usage and limit the effective context length.

\textbf{Effect of DINOv3 ViT size.} We further study how the choice of DINOv3 ViT affects overall performance. As shown in \cref{tab:dino_backbone}, using a larger ViT generally improves navigation success rate: ViT-B outperforms ViT-S by 4.2\%, and ViT-L brings a smaller additional gain to 59.4\%. However, this improvement comes at a substantial cost in model size: ViT-L has more than 3$\times$ the parameters of ViT-B. Considering both accuracy and efficiency, we adopt DINOv3 ViT-B as the default backbone for AstraNav-Memory, and use ViT-S / ViT-L mainly as ablation settings to probe the impact of tokenizer capacity.

Combined with our category-wise analysis in Sec.4.4 and the feature visualizations in Fig. 3, these results suggest that simply scaling the backbone mainly boosts performance on salient, object-centric targets (e.g., freezers or books), while texture- and boundary-sensitive categories (e.g., carpets) remain challenging. In other words, the remaining gap is less about backbone size and more about the type of visual cues available, motivating the integration of complementary boundary- or mask-level information in future work.

\begin{figure}[ht]
    \centering
    \includegraphics[width=1\linewidth]{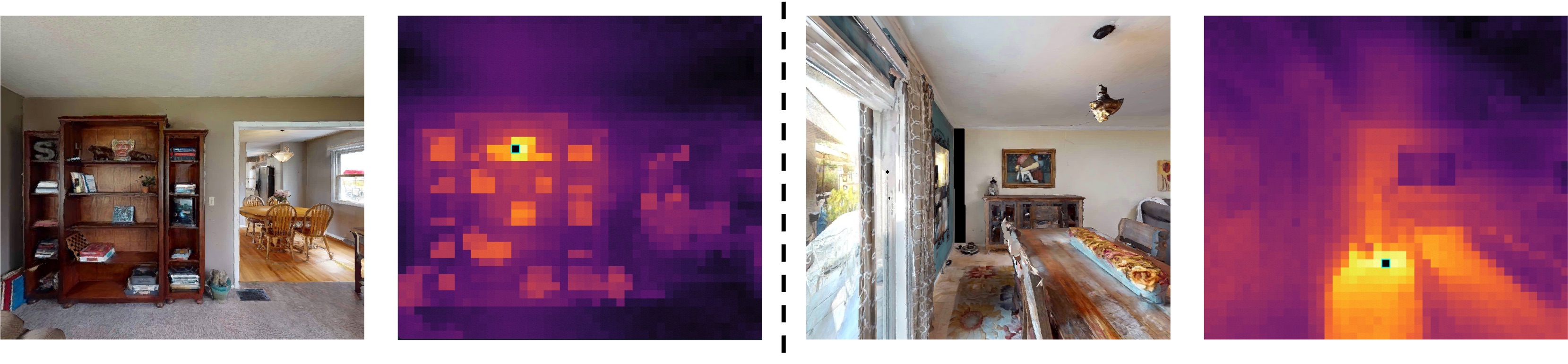}
    \caption{Visualization of DINOv3 patch features. For each image, we select a query patch (black square) and compute the similarity between its feature and all other patches, shown as a heatmap. Warmer colors indicate higher similarity to the selected patch.}
    \label{fig:vis_2}
\end{figure}

\subsection{Analysis}
We analyze the success rate by navigation target to understand how the two visual pipelines behave. The original model directly feeds all images into Qwen ViT, while AstraNav-Memory first encodes each image with DINOv3 ViT and then applies our token compression. As shown in \cref{tab:analysis}, we observe that AstraNav-Memory outperforms the original model on several categories that require recognizing relatively salient objects, such as freezer, piano, book, hanging clothes, and shower glass. In contrast, the original Qwen-only encoder works better on island, microwave, and especially carpet, suggesting that DINOv3 struggles to disentangle some fine-grained texture cues.

To better understand these category-wise differences, we further visualize patch-level features on representative examples. \cref{fig:vis_2} provides a qualitative analysis of these behaviors. For each image, we select a query patch (marked by a black square) and visualize the cosine similarity between its embedding and all other patches as a heatmap, where brighter regions indicate higher similarity. In the left part, we query a patch on a bookshelf: the heatmap correctly highlights other books on the shelf, showing that DINOv3 learns a coherent "book" concept, which explains why AstraNav-Memory achieves higher success rates on book targets. However, in the right part, when the query patch lies on a carpet, the response spreads across both the carpet and the surrounding floor, indicating that DINOv3 fails to clearly separate the two textures. This ambiguity is consistent with the lower performance of AstraNav-Memory on carpet compared to the original model. It also suggests that capturing such boundary-sensitive categories may require additional boundary- or mask-level cues—for example, by integrating features from segmentation models such as SAM into the compressed visual representation.



\begin{table}[t]
\centering
  \caption{Success rate comparison of models on different navigation targets.}
  \label{tab:analysis}
\begin{tabular}{c|cc}
\toprule
\multirow{2}{*}{Category} & \multicolumn{2}{c}{Model}    \\ \cline{2-3} 
                                                                   & Original & AstraNav \\ \hline
island                                                             & \textbf{97}       & 76.3              \\
microwave                                                          & \textbf{96}       & 75                \\
carpet                                                             & \textbf{61}       & 56                \\
freezer                                                            & 72                & \textbf{88}       \\
piano                                                              & 67                & \textbf{79}       \\
book                                                               & 66                & \textbf{73}       \\
hanging clothes                                                    & 63                & \textbf{80}       \\

shower glass                                                       & 53                & \textbf{92}       \\ 

\bottomrule
\end{tabular}
\end{table}

\begin{figure*}[h] 
\centering 
\includegraphics[width=1.0\linewidth]{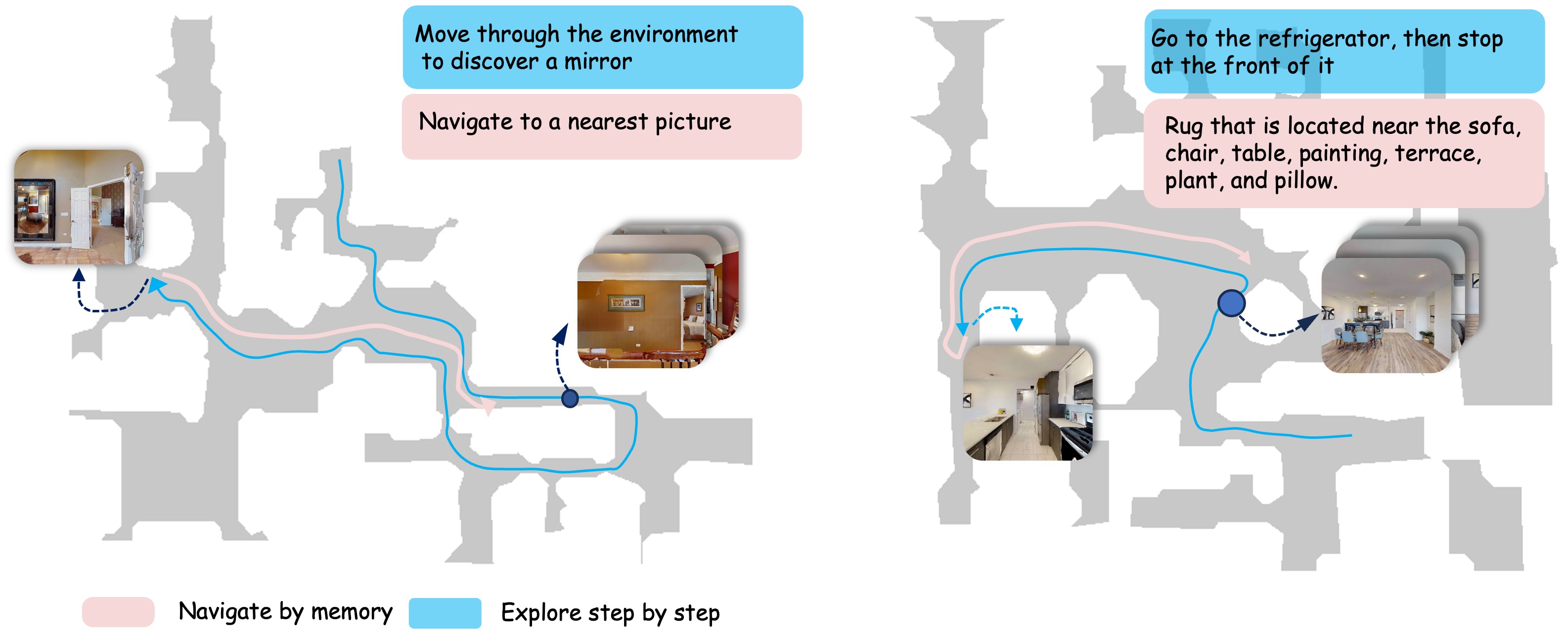}
\caption{Visualization of GOAT-Benchmark in Habitat-Sim, showing two different navigation modes: step-by-step exploration and optimal-path navigation based on memory.}
\label{fig:vis} 
\end{figure*}

\subsection{Visualization}
In \cref{fig:vis}, we present qualitative trajectories produced by our agent in novel environments. Each panel shows two tasks executed sequentially in the same scene. The blue curves denote the incremental exploration phase, during which the agent, guided by an initial generic instruction, gradually discovers objects (e.g., mirrors, pictures, refrigerators, rugs) and accumulates them into its image-centric memory. Once another language instruction is issued, the agent can utilize this memory and plans a direct route to the target, shown in pink. On the left case, after a short exploration, the agent can efficiently navigate to the nearest picture and then continue to a mirror without exhaustively scanning the house again. On the right case, the agent recalls the previously observed refrigerator and surrounding furniture and moves to the correct location of the rug with a short, near-optimal path. These visualizations highlight that our memory mechanism enables efficient, multi-goal navigation by reusing experience across tasks.


\section{Conclusion}

We introduced AstraNav-Memory, an image-centric memory framework for embodied lifelong navigation built on Qwen2.5-VL-3B. Instead of explicit maps or object queries, our model maintains long visual histories directly in the model context. A ViT-native visual tokenizer is constructed from frozen DINOv3 features and augmented with lightweight PixelUnshuffle and convolutional blocks. With 2$\times$ spatial downsampling, it compresses native vision tokens by 16$\times$, representing each frame with only 30 tokens while keeping the remaining vision–language stack unchanged. This makes long-horizon visual histories computationally affordable and tightly couples perception, language, and decision-making in an end-to-end manner. In practice, this allows the agent to retain hundreds of frames within a single context window without modifying downstream policy heads. Extensive experiments on GOAT-Bench and HM3D-OVON, together with ablations on memory length and compression rate, show that AstraNav-Memory achieves state-of-the-art navigation performance, with moderate compression providing the best balance between efficiency and accuracy. Category-wise analysis and feature visualizations show that our compressed representation captures salient object categories well but still struggles on boundary-sensitive targets, implying that adding boundary- and mask-aware cues is a promising field for future work.

\clearpage
\newpage
\bibliographystyle{assets/plainnat}
\bibliography{paper}

\end{document}